\documentclass[11pt]{article} 
\usepackage{rldmsubmit,palatino}
\usepackage{graphicx}
\usepackage{caption}
\usepackage{subcaption}
\usepackage{amsmath}
\usepackage{amssymb}
\usepackage{url}
\usepackage{multirow}
\usepackage[numbers]{natbib}
\title{Individual differences in the cognitive mechanisms of planning strategy discovery}

\author{
Ruiqi He\\
Max Planck Institute for Intelligent Systems\\
Tuebingen, BW 72076\\
\texttt{ruiqi.he@tuebingen.mpg.de} \\
\And
Falk Lieder \\
University of California, Los Angeles \\
Los Angeles, CA 90095 \\
\texttt{falk.lieder@psych.ucla.edu} \\
}

\begin{document}

\maketitle

\begin{abstract}
People employ efficient planning strategies.
But how are these strategies acquired? 
Previous research suggests that people can discover new planning strategies through learning from reinforcements, a process known as metacognitive reinforcement learning (MCRL). 
While prior work has shown that MCRL models can learn new planning strategies and explain more participants' experience-driven discovery better than alternative mechanisms, it also revealed significant individual differences in metacognitive learning. 
Furthermore, when fitted to human data, these models exhibit a slower rate of strategy discovery than humans.
In this study, we investigate whether incorporating cognitive mechanisms that might facilitate human strategy discovery can bring models of MCRL closer to human performance. 
Specifically, we consider intrinsically generated metacognitive pseudo-rewards, subjective effort valuation, and termination deliberation. 
Analysis of planning task data shows that a larger proportion of participants used at least one of these mechanisms, with significant individual differences in their usage and varying impacts on strategy discovery. 
Metacognitive pseudo-rewards, subjective effort valuation, and learning the value of acting without further planning were found to facilitate strategy discovery. 
While these enhancements provided valuable insights into individual differences and the effect of these mechanisms on strategy discovery, they did not fully close the gap between model and human performance, prompting further exploration of additional factors that people might use to discover new planning strategies. 
\end{abstract}



\startmain 

\section{Introduction}
Planning is a fundamental aspect of daily life, whether it involves preparing the next meal or laying out long-term career goals. 
This process of planning can be represented as a search tree, which expands rapidly with the number of possible actions and the length of the planning horizon. 
To manage this complexity, computers can rely on more computational resources or advanced algorithms designed to enable efficient search across the decision tree. 
Similarly, people use sophisticated strategies to plan effectively \cite{callaway2022rational}.
However, where do these strategies originate? 
While previous work shed some light on how people learn to select between existing strategies \cite{rieskamp2006ssl, LiederGriffiths2017} and how children acquire arithmetic strategies \cite{siegler1991microgenetic}, less is known about the mechanisms behind the discovery of planning strategies. 
Recent work by \citet{he2024experiencedrivendiscoveryplanningstrategies} proposed that planning strategies emerge from learning through experience, a concept that falls under the broader idea of \textit{metacognitive reinforcement learning (MCRL)}. 
Their work introduced MCRL models that are capable of learning new planning strategies and demonstrated (i) that MCRL models better account for the experience-driven discovery of more participants compared to alternative learning mechanisms; and (ii) that there are individual differences in metacognitive learning. 
When fitting the models' hyperparameters to participants, the models, however, fell short of discovering the new planning strategy at the same rate as participants did. 

In this study, we build upon one of \citet{he2024experiencedrivendiscoveryplanningstrategies}'s MCRL models by enhancing it with additional cognitive mechanisms to investigate whether we can bridge this performance gap as well as explore the extent and role of individual differences in these mechanisms, examining how they may affect strategy discovery. 
The following sections introduce \citet{he2024experiencedrivendiscoveryplanningstrategies}'s experiment on strategy discovery, outline the concept of MCRL, describe the additional cognitive mechanisms as model extensions, and conclude with the modeling results. 

\section{Experiment}
To examine the additional cognitive mechanism behind strategy discovery, we used data from the experiment by \citet{he2024experiencedrivendiscoveryplanningstrategies} where participants solved a series of planning tasks that all shared the same unique, adaptive planning strategy, which participants needed to discover as it was not part of their prior mental repertoire.
Their analysis showed that participants indeed exhibited significant evidence of experience-driven strategy discovery: the proportion of adaptive strategies increased from 0.79\% (CI: [0\%; 2.22\%]) in the first trial to 28.57\% (CI: [27.21\%; 29.93\%]) after 120 trials.

\section{Modelling experience-driven strategy discovery}
To model this observed experience-driven discovery, we built on \citet{he2024experiencedrivendiscoveryplanningstrategies}' metacognitive hybrid Reinforce model, which offered a superior explanation of how people learn to adapt their amount of planning and their planning strategies \cite{he2023mechanisms}. 
Before introducing our additional cognitive mechanisms, we will first explain the concept of MCRL.

\subsection{From reinforcement learning to metacognitive reinforcement learning}
Reinforcement learning (RL) has emerged as a powerful framework for modeling how humans learn from interactions with their environment. 
Similar to human trial-and-error learning, a common form of RL involves predicting the expected reward, known as Q-value $Q(s,a)$, for taking a specific action $a$ in a given state $s$, based on previous experiences and feedback \cite{watkins1992q}. This is formalized through the following update rule for the Q-value:
$Q(s,a) \leftarrow Q(s,a) + \alpha \cdot \delta$,
where $\alpha$ is the learning rate, and $\delta$ is the reward prediction error, which is the difference between actual and predicted reward.

While traditional reinforcement learning focuses on modeling external actions, metacognitive reinforcement learning focuses on internal mental operations that govern decisions about how to think or decide, a concept referred to as meta-decision-making \cite{Boureau2015}. 
These processes are formalized as a meta-level MDP \cite{hay2014selecting}, represented as:
$M_{meta}=\left( \mathcal{B}, \mathcal{C} \cup \lbrace \bot \rbrace, \mathcal{T}_{meta}, \mathcal{R}_{meta} \right)$. 
Here, $b_t \in \mathcal{B}$ denotes the mental belief state at time $t$, $c_t\in \mathcal{C}$ the mental computations including deliberation termination $\bot$, $\mathcal{T}_{meta}$ the transitions between mental belief states, and $\mathcal{R}_{meta}$ the trade-off between the cost of computations and the expected return from terminating deliberation and acting based on the current belief state.
Since solving a meta-MDP is often computationally intractable, the brain is hypothesized to approximate optimal meta-decision-making through reinforcement learning \cite{callaway2018learning}. 

\subsection{Strategy representation}
Before describing the MCRL model, we will first outline how planning strategies are represented. 
In the MCRL model, planning strategies are represented by the weights of 63 features of potential belief-computation pairs. 
Details about these features are available at \url{https://osf.io/6wxyh/?view_only=0da5edc1bc794f4085bb745917173f9d}. 
This representation captures an individual’s learning trajectory as a combination of features and their feature weights that evolve over trials through the learning process.

\subsection{Gradient ascent through the strategy space: Reinforce}
The metacognitive Reinforce model \cite{jain2019measuring}, inspired by the Reinforce model \cite{Williams1992}, posits that individuals refine their planning strategies within a defined space of possible strategies. 
These strategies, akin to the commonly used term of policy in the context of policy-gradient reinforcement learning methods, map features of belief states to values of cognitive operations, which are converted into probabilities for specific planning operations using a softmax function \cite{Williams1992}.
Concretely, the strategy is parameterized by a weight vector $\mathbf{w}$, which is updated after each trial to account for the effectiveness of different cognitive operations:
$ \mathbf{w} \leftarrow \mathbf{w} + \alpha \cdot \sum_{t=1}^{O}\gamma^{t-1} \cdot r_{meta}(b_t,c_t) \cdot \nabla_\mathbf{w} \ln \pi_\mathbf{w}(c_t | b_t), 
$
where $O$ is the number of planning operations performed during the trial, $b$ represents the belief state, $c$ is the cognitive operation under evaluation, $\mathcal{C}_{b}$ is the set of available cognitive operations in state $b$, $\alpha$ is the learning rate, and $\gamma$ is the discount factor.
The learned weight vector, combined with the previously described features, is used to approximate the meta-level Q-values as:
 $Q_{\text{meta}}(b_t,c_t) \approx \sum_{j=1}^{56} w_j \cdot f_j(b_t,c_t) $
These Q-values are then used to select cognitive operations probabilistically, through the softmax function
$P(c_t|b_t,Q_{\text{meta}}) \propto \exp(Q_{\text{meta}}(b_t,c_t) / \tau)$.
The model has the free hyperparameters $\alpha$, $\gamma$, the inverse temperature $\tau$, and the 63 initial feature weights.

While \citet{he2024experiencedrivendiscoveryplanningstrategies} demonstrated that the metacognitive Reinforce is capable of discovering the new planning strategy, when simulating model performance with free hyperparameters fitted to human discovery data,  the models adapt more slowly than participants. 
A possible explanation is that people rely on additional cognitive mechanisms to aid strategy discovery.
To test this, we enhanced the Reinforce model with these additional mechanisms, which we will introduce next.

\subsection{Additional cognitive mechanisms}
We augmented the plain metacognitive Reinforce with three additional components: intrinsic rewards for generating valuable information (pseudo-rewards), imposing a subjective value of effort (SE), and deliberation about the value of terminating planning (termination deliberation).

\paragraph{Pseudo-reward (PR)}
The central role of reward prediction errors in reinforcement learning, combined with the scarcity of external rewards in metacognitive learning, suggests that the brain may accelerate learning by generating metacognitive pseudo-rewards (PR). 
Supporting this, prior work \cite{srinivas2023learning} showed that people use intrinsic self-evaluative signals when external feedback is limited. 
Building on these insights, PR was added to the metacognitive Reinforce model to quantify the value of information from recent planning operations.
Specifically, the PR for transitioning from one belief state, $b_t$, to the next, $b_{t+1}$, is the difference in the expected value of the agent’s optimal path before and after the transition, reflecting how much the new belief state improves upon the previous one: $
\text{PR}(b_t,c,b_{t+1})= \mathbb{E}[R_{\pi_{b_{t+1}}}|b_{t+1}]-\mathbb{E}[R_{\pi_{b_t}}|b_{t+1}] $
where $\pi_b(s) = \text{\text{argmax}}_a \mathbb{E}_b[R \mid s, a]$  represents the policy that the agent would follow given belief state $b$, and $R$ denotes the expected cumulative external reward based on the probability distribution $b$.
This PR is added to the reward signal that the metacognitive Reinforce model learns from. 
The PR mechanism incorporates one additional free hyperparameter, the PR weight, which modulates its impact on the learning process \cite{he2023mechanisms}.

\paragraph{Subjective value of effort (SE)}
People vary in how they perceive the effort of processing information, with some experiencing it as more unpleasant and are therefore more effort-avoidant, while others are effort-seeking because they perceive the additional effort worthwhile \cite{shenhav2017toward}. 
This continuous variation in the (un)pleasantness of effort was not accounted for in the plain Reinforce model, which imposed a fixed cost for gathering information uniformly across all participants. 
Therefore, we enhanced the model by adding one additional free hyperparameter, referred to as the subjective value of effort (SE) \cite{he2023mechanisms}, which is added to the reward signal of planning operations.

\paragraph{Termination deliberation (TD)}
The value of termination can either be learned like the other planning operations or, alternatively, people could engage in metareasoning \cite{Griffiths2019} to determine the expected value of terminating planning based on the current belief state. 
This is akin to asking, ``What is my estimated reward if I stop planning now and start acting based on my current plan?'' and assigning this value to the termination action.
Termination deliberation (TD) is therefore implemented by calculating the expected return at the current belief state and using it as the $Q_{meta}$-value of the termination action \cite{he2023mechanisms} after every planning operation.

Considering all possible combinations of these extensions resulted in 8 model variants: plain Reinforce, Reinforce with PR, Reinforce with TD, Reinforce with SE, Reinforce with PR and TD, Reinforce with PR and SE, Reinforce with SE and TD, Reinforce with PR, SE and TD. 

\subsection{Model fitting and model selection}
To evaluate how well the model variants can explain human planning, we fitted each one to each participant's click sequences by maximizing their likelihoods through Bayesian optimization \cite{bergstra2013making} using 60,000 iterations.
The likelihood of a click sequence is calculated as the product of the probabilities assigned to each individual click.
After model fitting, we conducted family-level Bayesian model selection (BMS) by categorizing the variants based on their mechanisms: PR vs. no PR variants, TD vs. no TD variants, and SE vs. no SE variants. 
Specifically, we estimated the expected proportion of participants whose behavior is best captured by each model family ($r$) and computed the exceedance probability ($\phi$), which quantifies the confidence that a given family is favored over the other using random-effects Bayesian model selection. 
In addition, we evaluated the relative fit of each model to individual participants' data using the Bayesian Information Criterion (BIC). 

\section{Modeling results}
\begin{figure}[ht]
\centering
\begin{subfigure}[]{0.49\textwidth}
\centering
\begin{tabular}{ccc}
\hline
Variant group & $r$ & $\phi$ \\ \hline
PR & 29.89\% & 0\% \\
No PR & 70.11\% & 100\% \\
TD & 9.39\% & 0\% \\
No TD & 90.61\% & 100\% \\
SE & 34.75\% & 0\% \\
No SE & 65.25\% & 100\% \\ \hline
\end{tabular}
\caption{BMS results where $r$ described the proportion of participants best explained by a model variant family and $\phi$ the probability that this proportion is higher than the alternative family. Note that participants could be classified into more than one variant group, as classification is based on the presence of model elements rather than mutually exclusive categories. For example, one participant could be best explained by a variant that includes both PR and SE (but no TD), thus overlapping both PR and SE groups.}
\label{table:exp3variantsall}
\end{subfigure}
\hspace*{\fill}
\begin{subfigure}[]{0.49\textwidth}
\centering
\includegraphics[width=0.9\linewidth]{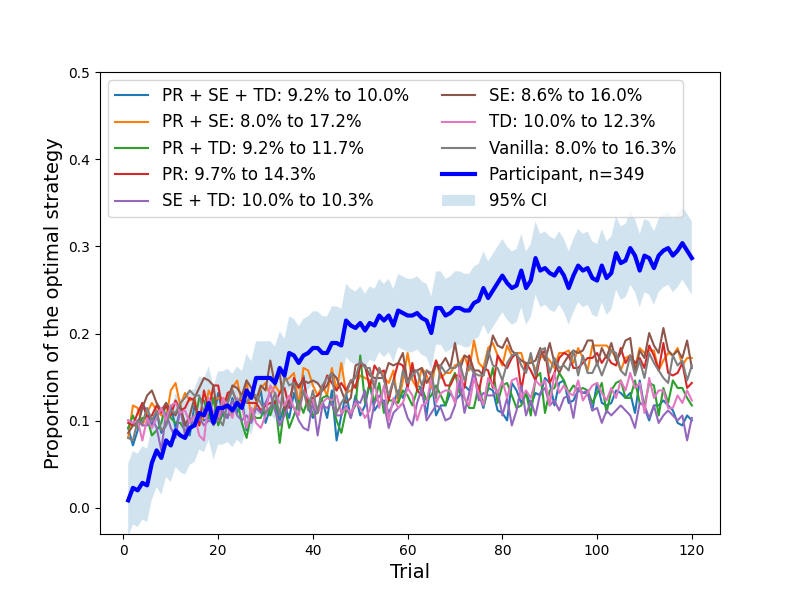}
\caption{Simulated proportion of the adaptive strategy of the participants and of the model variants.}
\label{fig:exp3modelvariantssimulation}
\end{subfigure}
\hspace*{\fill}
\caption{Comparison of BMS results (left) and simulated proportions (right) of the Reinforce model variants.}
\label{fig:1}
\end{figure}

Overall, the model variants provided a better relative fit for 244 participants (69.96\%) compared to the plain Reinforce (102; 30.09\%). 
Looking at the mechanism individually revealed substantial differences: not everyone used the same cognitive mechanisms. There was notable variation in which variant explained a participants' data best (see Figure~\ref{table:exp3variantsall}).
To investigate the implications of these individual differences, participants were grouped based on the best relative model fit, that is, the variant with the lowest BIC. 
Analysis revealed significant performance discrepancies across these groups. 
Participants whose behavior was best explained by model variants incorporating PR (30\% of the participants; referred to as PR participants) or SE (35\% of the participants; referred to as SE participants) scored significantly higher than those whose behavior was fitted by models without these mechanisms, while participants engaging in TD (9\% of the participants; referred to as TD participants) performed worse than those who learned the value of termination like the other planning operations
(see Table \ref{table:exp3variantsindividualdifference}).  
When examining the amount of planning, PR and SE participants performed significantly more planning operations, whereas the opposite was true for those who deliberated about termination (see Table \ref{table:exp3variantsindividualdifferenceclicks}). 
PR and SE participants' number of planning operations were closer to the average number of mental operations associated with the optimal planning strategy (1.67 operations).
\begin{table}[ht]
\centering
\begin{subtable}[t]{0.49\textwidth}
\centering
\begin{tabular}{ccccc}
\hline
Variant family & Mean & Std & Test statistic & $p$-value \\ \hline
PR & -2.99 & 34.22 & \multirow{2}{*}{56203259} & \multirow{2}{*}{\textless{}.001} \\
No PR & -4.69 & 34.15 &  &  \\ \hline
TD & -7.79 & 34.99 & \multirow{2}{*}{38432219} & \multirow{2}{*}{\textless{}.001} \\
No TD & -2.48 & 33.79 &  &  \\ \hline
SE & -2.97 & 34.83 & \multirow{2}{*}{57113923.5} & \multirow{2}{*}{\textless{}.001} \\
No SE & -4.81 & 33.50 &  &  \\ \hline
\end{tabular}
\caption{Mean and std of the score of participants best fitted by the respective model variant and the Mann Whitney U test results.}
\label{table:exp3variantsindividualdifference}
\end{subtable}
\hfill
\begin{subtable}[t]{0.49\textwidth}
\centering
\begin{tabular}{ccccc}
\hline
Variant family & Mean & Std & Test statistic & $p$-value \\ \hline
PR & 1.87 & 1.35 & \multirow{2}{*}{59095962.5} & \multirow{2}{*}{\textless{}.001} \\
No PR & 1.25 & 1.35 &  &  \\ \hline
TD & 0.43 & 0.91 & \multirow{2}{*}{16464819} & \multirow{2}{*}{\textless{}.001} \\
No TD & 1.70 & 1.36 &  &  \\ \hline
SE & 1.87 & 1.32 & \multirow{2}{*}{63198092} & \multirow{2}{*}{\textless{}.001} \\
No SE & 1.22 & 1.36 &  &  \\ \hline
\end{tabular}
\caption{Mean and std of the planning amount measured by the number of clicks performed by each participant group and the Mann Whitney U test results.}
\label{table:exp3variantsindividualdifferenceclicks}
\end{subtable}
\caption{Comparison of score (left) and amount of planning (right) between participant groups.}
\end{table}

To evaluate whether the additional mechanisms improved the models' ability to align more closely with human behavior, we fitted each model variant's free hyperparameters to each participant's click sequences and analyzed the simulated planning. 
Pairwise comparisons of the proportion of simulated participants using the adaptive planning strategy between each model variant and the plain Reinforce model revealed that none of the variants achieved significantly closer alignment with participants' rate of discovery than the plain model, indicating that the performance gap identified by \citet{he2024experiencedrivendiscoveryplanningstrategies} still persists despite the additional cognitive mechanisms included in the new model variants (all $p<.001$; see Figure~\ref{fig:exp3modelvariantssimulation} for the simulations).
Conducting the same analysis only for the participants best explained by each model variant resulted in similar results: all model variants fall short of discovering the adaptive strategy at the same rate as the participants. 
Details on the regression analysis can be found at \url{https://osf.io/uj2wt}. 


\section{Discussion}
To investigate the mechanism behind strategy discovery, \citet{he2024experiencedrivendiscoveryplanningstrategies} introduced metacognitive reinforcement learning (MCRL) models and tested those models on data collected from a planning task that required learning a unique planning strategy. 
While their MCRL models explained the majority of participants' behavior better than alternative models, two key issues were identified: 1) individual differences in metacognitive learning mechanisms, which led to variations in performance, and 2) a performance gap between the models when fitted to participants’ data compared to actual participant performance.
To address this gap and explore individual differences in relation to additional cognitive mechanisms that people might use to facilitate strategy discovery, we extended the most promising MCRL model by incorporating pseudo-reward (PR), subjective value of effort (SE), and termination deliberation (TD). 
Overall, 70\% participants used at least one of the additional mechanisms, with substantial differences among the individuals. 
Participants better described by a variant incorporating PR and SE performed significantly better, though only about 30\% and 35\% of participants were best explained by variants using these mechanisms. 
This suggests that while PR and SE can enhance performance, they may demand additional cognitive effort, making people hesitant to use them unless the benefits justify the cost.
Interestingly, participants who engaged in TD tended to terminate their planning early, leading to poorer performance.
Although the additional cognitive mechanisms did not significantly enhance the model’s alignment to human discovery performance, they provided valuable insights into individual differences and the role of these mechanisms in metacognitive learning, contributing to a deeper understanding of human metacognitive learning.


\bibliographystyle{abbrvnat}
{\footnotesize\bibliography{rldm.bib}}
\end{document}